# Expression Analysis Based on Face Regions in the Wild

Zheng Lian[1,2]   Ya Li[1]   Jianhua Tao[1,2,3]   Jian Huang[1,2]   Mingyue Niu[1,2]

[1]National Laboratory of Pattern Recognition, Institute of Automation Chinese Academy of Sciences, Beijing, PRC

[2] School of Artificial Intelligence, University of Chinese Academy of Sciences, Beijing, PRC

[3] CAS Center for Excellence in Brain Science and Intelligence Technology, Institute of Automation Chinese Academy of Sciences, Beijing, PRC

**Abstract:** In the field of human-machine interaction, facial emotion recognition is an essential and important aspect. Past research on facial emotion recognition focuses on the laboratory environment. However, it faces many challenges in the real-world conditions, i.e., illumination changes, large pose variations and partial or full occlusions. Those challenges cause different face areas with different sharpness and completeness. Inspired by this fact, we focus on the authenticity of predictions generated by different <emotion, region> pairs. For example, if only the mouth areas are available and the emotion classifier predicts happiness, then how to judge the authenticity of predictions. This problem can be converted into the contribution of different face areas to different emotions. In this paper, we divide the whole faces into six areas, including nose areas, mouth areas, eyes areas, nose to mouth areas, nose to eyes areas and mouth to eyes areas. To obtain more convincing results, our experiments are conducted on three different databases: FER+, RAF-DB and ExpW. Through analysis on the classification accuracy, the confusion matrix and the Class Activation Map (CAM), we can conclude convincing results. To sum up, contributions of this paper lie in two aspects: (1) We visualize concerned areas of human in emotion recognition; (2) We analyze the contribution of different face areas to different emotions in the wild through experimental analysis. Our findings can be combined with psychological findings to promote the understanding of the emotion expression.

## 1   Introduction

With the development of artificial intelligence, there is an explosion of interest in realizing more natural human-machine interaction (HMI) systems. Inspired by psychological findings, Prendinger et al. [1] and Martinovski et al. [2] point out that addressing emotion information in the conversation agents or the dialogue systems can enhance satisfaction and cause fewer breakdowns in the dialogue. Therefore, emotion recognition, as an essential aspect in HMI, is attracting more and more attention [3-5].

In the field of emotion recognition, facial expression recognition is a hot research topic due to its wild applications. For example, there are millions of images are being uploaded every day by different users. Their emotion states are useful for recommendation systems to determine whether to push product information. To automatic recognize the affective state of face images from the Internet, facial expression recognition is essential.

Past research on facial expression recognition is a multi-step process, where handicraft features are extracted first, combined with various classifiers and fusion methods. In general, facial features consist of two parts: appearance features and geometry features. As for appearance features, Histogram of Oriented Gradient (HOG) [6], Local Binary Patterns (LBP) [7], Local Phase Quantization (LPQ) [8] and Scale Invariant Feature Transform (SIFT) [9] are wildly utilized. As for geometry features, the head pose and landmarks are also considered in emotion recognition.

However, targets of the multi-step approach are not consistent. Besides, there is no agreement on appropriate handicraft features for emotion recognition. To solve these problems properly, the multi-step approach is replaced by the end-to-end method, which has gained state-of-the-art performance in many tasks, such as image classification [10], machine translation [11], scene classification [12], image caption generation [13] and speech synthesis [14]. In the end-to-end facial emotion recognition system, original faces cropped into standard size are treated as inputs. And, emotional labels are treated as outputs. End-to-end image classifiers, including AlexNet [15], VGG [16], GoogLeNet [17], ResNet [18], DenseNet [19] and other variation of those models, are trained to map inputs to corresponding outputs.

Despite great efforts have been made to improve the performance of the facial expression recognition, many challenges still exist. In the real-world conditions, it's difficult to gather faces without the shade from other objects. In the meantime, faces are not always in the frontal pose and the proper light conditions. Therefore, front faces without any noise are not always available in the emotion recognition task.

This question can be partially solved by conducting emotion recognition based on partial faces. The pioneer work by Ekman el al. [20] proposed Facial Action Coding System (FACS), which described facial expression as the combination of multiple action units. Followed with [20], Tian et al. [21] focused on analyzing different facial parts, i.e., eyes, nose and mouth, and mapping them into Action Units (AU) coding. As for facial expression recognition, Wang et al. [22] combined FACS and LBP to represent facial expression features from coarse to fine. The facial feature regions were extracted by FACS, and then LBP was used to represent expression features for enhancing the discriminant. Sun et al. [23] proposed to recognize facial expression based on regions of interest, which guided convolutional neural networks (CNNs) to focus on areas associated with the expression. Zhang et al. [24] learned the joint representation by considering the texture and landmark modality of facial images. They divided faces into different patches, and then concatenated these corresponding patches as individual vectors. Zaman et al. [25] proposed a feature selection process to represent facial features contribution according to their variations. However, previous works did not analyze the contribution of different face areas for different emotions. Besides, they mainly focused on the lab-controlled environment.

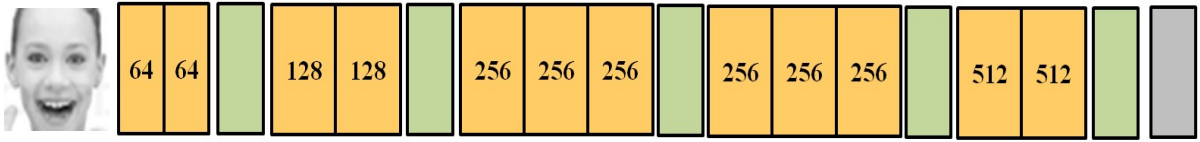

Fig. 1. Flowchart of our **classification** system. Yellow boxes, green boxes and grey boxes denote the 2D convolutional layers, max pooling layers and fully-connected layers, respectively. The number inside of the yellow box is the number of filters. And the number of neurons of the grey box is the same as the number of categories.

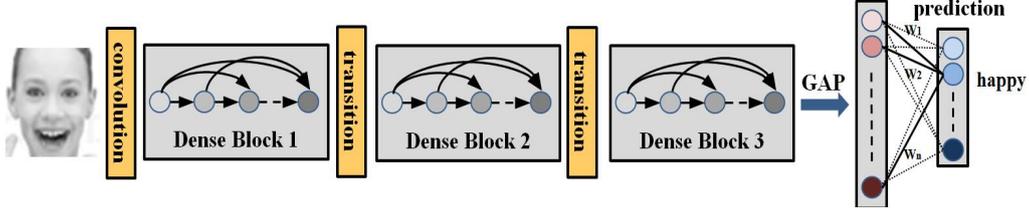

Fig. 2. Flowchart of our **visualization** system. Three dense blocks are followed behind the inputs, combining global average pooling (GAP) and a fully-connected layer in the end. The outputs of the system are normalized emotion probabilities.

Considering the limitation of previous works [20-25], in this paper, we focus on the authenticity of predictions generated by different <emotion, region> pairs in the wild. For example, if only the mouth areas are available and the emotion classifier predicts happiness, then how to calculate the confidence scores of predictions. This question can be converted into another question: how much information of happiness can be expressed through mouth areas. This question mainly focuses on psychological aspect, and few studies focus on it.

To solve this problem to some extent, we divide the whole faces into six areas, including nose areas, mouth areas, eyes areas, nose to mouth areas, nose to eyes areas and mouth to eyes areas. What's more, we analyze the contribution of different face areas to different emotions in the wild. And, we visualize concerned areas of human in emotion recognition through Class Activation Mapping (CAM) [26]. To obtain more convincing results, our experiments are conducted on three different databases: FER+, RAF-DB and ExpW. Our work has some similarities with Busso et al. [27]. They separate the whole face into the forehead, eyebrow, low eye, right check and left check, and then a separate classifier is implemented for each block. Their experiments are conducted in the lab-controlled environment. However, in our paper, experiments are conducted in real-world conditions. In the meantime, the whole face is divided into smaller parts and more evaluation approaches are adapted. It is reasonable to believe that our findings can promote the understanding of the emotion expression.

This paper is organized as follows. In Section 2, we describe the proposed system in detail. Experimental setup and evaluation results are illustrated in Section 3 and Section 4, respectively. In Section 5, we conclude the whole paper.

## 2 System description

In this section, the classification model and the visualization model are discussed in detail.
1. As for the classification model, we follow the VGG network.
2. As for the visualization model, we follow the Dense-Net-BC architecture in [19]. Through the CAM technique in the visualization model, we can visualize activation parts of different inputs.

Although the visualization model can also be utilized for the classification, we figure out that our classification model can obtain higher classification accuracy than the visualization model. Therefore, we split our classification process and visualization process into two parts. The architecture of two models can be found in Fig. 1 and Fig. 2, respectively.

### 2.1 Classification model

As for the classification model, we follow the VGG architecture, which consists of multiple convolutional layers, max pooling layers and a fully-connected layer (FC). The inputs of the system are grey-scale images in $64 \times 64$ pixels and the outputs are normalized emotion probabilities.

The system architecture is shown in Fig. 1. This network increases the depth using an architecture with very small ($3 \times 3$) convolution filters, whose convolutional stride is fixed to 1 pixel. Batch Normalization [28] and ReLU [29] are also added after the convolutional layers. Batch Normalization alleviates the gradient explosion problems and ReLU is chosen as the activation functions. The max-pooling layer is appended behind two or three convolutional layers, which is performed over a $2 \times 2$ pixel window with stride 2. After multiple convolutional layers and max-pooling layers, a FC layer is connected behind to generate emotion probabilities, whose output dimension is the same as the number of categories in the dataset.

## 2.2 Visualization model

In the visualization model, we follow the DenseNet-BC architecture, which has three dense blocks associated with the global average pooling (GAP) and the FC behind. The inputs of the system are 64×64 grey-scale images and the outputs are normalized emotion probabilities.

The system architecture is shown in Fig. 2. Before entering into the first dense block, the convolutional layer with 16 output channels is performed on the 64×64 grey-scale images. Three dense blocks are followed behind and each dense block has 16 layers. In each dense block, 3×3 convolutional filters are used combining zero-padding with one pixel to keep the feature-map size fixed. Batch Normalization is also added before convolutional layers to alleviate the gradient explosion problems. Between contiguous dense blocks, a transition block is applied to reduce the size and the channel of feature maps. The transition block is composed with a 1×1 convolutional layer, followed with 2×2 average pooling behind. Finally, the GAP and FC are combined to generate emotion probabilities.

## 2.3 CAM technique

CAM is adapted to visualize activation parts of different emotions, which projects back the weights of the output layer on to the convolutional feature maps to identify the importance of the image regions. Concretely, we utilize a weighted sum on the outputs of GAP, which are spatial average of the feature maps generated from the last dense block.

We formulate the process of GAP as:

$$F_k = \sum_{x,y} f_k(x, y) \quad (1)$$

where $f_k(x, y)$ represents value of $(x, y)$ in the $k^{th}$ feature map extracted from the last dense block. $F^k$ represents the $k^{th}$ output of GAP, which is spatial average of the $k^{th}$ feature map.

To class $c$, the class score:

$$\begin{aligned} S_c &= \sum_k w_k^c F_k \\ &= \sum_k w_k^c \sum_{x,y} f_k(x,y) \\ &= \sum_{x,y} \left( \sum_k w_k^c f_k(x,y) \right) \end{aligned} \quad (2)$$

where $w_k^c$ represents the value connected the $k^{th}$ output of GAP to the class $c$.

Therefore, it is the CAM for inputs, which directly indicates importance of activation at spatial coordinate $(x, y)$ related to class $c$. By upsampling the CAM to the size of inputs, we can identify image regions, which are the most relevant to the particular category.

# 3 Experimental setup

Our system is tested on the FER+ dataset [30], RAF-DB dataset [31] and the ExpW dataset [32]. These datasets contain seven or eight emotion categories. Seven basic emotion categories, including neutral, happiness, surprise, sadness, anger, disgust and fear, are all contained in three datasets except that contempt is also considered in the FER+ dataset. These datasets are collected in the wild and their emotions are more natural than existing databases [33-37]. In the meantime, their quantity is sufficient to train a robust deep network.

To divide the whole faces into different facial parts, the open-source library, Dlib [38], is also utilized.

## 3.1 FER+ database

The FER+ database is an extension of the FER database [39]. They re-label each image in the FER database through ten crowd taggers to overcome the noise label issue.

The FER dataset is created to mimic real-world conditions through Google image search API. It consists of 35887 images: 28709 for the training, 3589 for the public testing and 3589 for the private testing. The dataset consists of 48×48 pixel grey-scale facial images. Each face is more or less centered and occupies about the same amount of space. The task is to categorize each face based on the emotion shown in the facial expression into one of seven categories, including neutral, happiness, surprise, sadness, anger, disgust and fear.

Compared with FER, FER+ has eight emotion categories adding contempt as well. We follow the same data selection method provided in [30]. If less than 50% of the votes are integrated, the sample will be removed. Then we combine the training data and the public testing set as the training set and evaluate the model performance on the private testing set. Data distribution of the training set and the testing set is shown in Table 1.

Table 1 Class category distribution of the FER+ dataset.

|  | Train | Test | Total |
|---|---|---|---|
| Neutral | 11000 | 1219 | 12219 |
| Happiness | 8326 | 920 | 9246 |
| Surprise | 3807 | 429 | 4236 |
| Sadness | 3660 | 421 | 4081 |
| Anger | 2535 | 287 | 2822 |
| Disgust | 151 | 19 | 170 |
| Fear | 636 | 88 | 724 |
| Contempt | 153 | 21 | 174 |
| Sum | 30268 | 3404 | 33672 |

## 3.2 RAF-DB database

RAF-DB database is a real-world expression database, which contains 29672 real-word facial images collected by Flickr's image search API. They employed 315 annotators who have been instructed with one-hour tutorial on emotion for an online facial expression annotation assignment. Finally, each image was labeled by about 40 independent labelers. Subjects in the RAF-DB database are range from 0 to 70 years old. There are 52% female, 43% male and 5% remains unsure.

RAF-DB database is divided into single-label subset and two-tab subset. The single-label subset contains seven classes of basic emotion, including neutral, happiness, surprise, sadness, anger, disgust and fear, and the two-tab subset contains twelve classes of compound emotions. In the experiment, we follow the same data selection method provided in RAF-DB and only utilize the single-label subset, which contains 15339 images: 12271 for the training and 3068 for the testing. Data distribution of the training set and the testing set is shown in Table 2.

Table 2　Class category distribution of the RAF-DB dataset.

|  | Train | Test | Total |
|---|---|---|---|
| Neutral | 2524 | 680 | 3204 |
| Happiness | 4772 | 1185 | 5957 |
| Surprise | 1290 | 329 | 1619 |
| Sadness | 1982 | 478 | 2460 |
| Anger | 705 | 162 | 867 |
| Disgust | 717 | 160 | 877 |
| Fear | 281 | 74 | 355 |
| Sum | 12271 | 3068 | 15339 |

### 3.3　ExpW database

ExpW database is a real-world expression database, which contains 91793 real-word facial images manually labeled with expressions. Each image in the ExpW dataset is labeled into one of seven basic emotion categories: neutral, happiness, surprise, sadness, anger, disgust and fear.

Images in ExpW dataset are collected by Google image search API. At first, they combine a list of emotion-related keywords with different nouns as queries for Google image search. Then they collect images returned from the search engine and run a face detector [40]. Non-face images are removed. Images in the ExpW dataset have larger quantity and more diverse face variations than many databases.

The face confidence score is provided for each image in ExpW, which is range from 0 to 100. To analyze on a cleaner subset, we only choose face image whose confidence score is greater than 60 in the experiment. Since there is no existing separation approach for the training set and testing set, we split the dataset into the training set and testing set by a ratio of 4:1 while keeping the original label distribution. In the end, we utilize 33374 images in the ExpW for the experiment: 26701 for training and 6673 for testing. Data distribution of the training set and the testing set is shown in Table 3.

Table 3　Class category distribution of the ExpW dataset.

|  | Train | Test | Total |
|---|---|---|---|
| Neutral | 8309 | 2077 | 10386 |
| Happiness | 10576 | 2644 | 13220 |
| Surprise | 2471 | 617 | 3088 |
| Sadness | 2494 | 623 | 3117 |
| Anger | 1272 | 318 | 1590 |
| Disgust | 1250 | 312 | 1562 |
| Fear | 329 | 82 | 411 |
| Sum | 26701 | 6673 | 33374 |

### 3.4　Face region extraction

To divide facial images into different parts, facial landmark detection is essential. We utilize the open-source library, Dlib library, to detect landmarks, which takes the now classic HOG feature set combined with a linear classifier, an image pyramid and the sliding window detection scheme [41]. 68 landmarks are detected by Dlib, which can be found in Fig. 3.

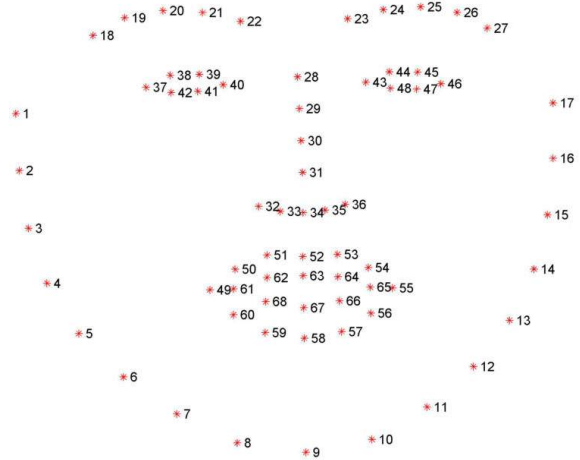

Fig.3. 68 landmarks that are detected by the Dlib library.

After landmark extraction, we divide the whole faces into six face regions based on the position of corresponding landmarks, including nose areas, mouth areas, eyes areas, nose to mouth areas, nose to eyes areas and mouth to eyes areas. In the division process, we utilize a 'mask' to contain all landmarks of corresponding regions. Landmarks that should be contained for each region can be found in Table 4, whose index is the same as the index in Fig. 3. For example, the mouth region should contain $49^{th}$~$68^{th}$ landmarks.

Table 4　Landmarks for each region.

| Face areas | Landmark index |
|---|---|
| Mouth | [49, 68] |
| Nose | [29, 36] |
| Eyes | [18, 22] + [37, 42] |
| Nose and mouth | [29, 36]+ [49, 68] |
| Nose and eyes | [18, 22] + [29, 36] + [37, 42] |
| Mouth and eyes | [18, 22] + [49, 68] + [37, 42] |
| The whole faces | [1, 68] |

Finally, an example of the division process is shown in Fig. 4. Through analysis on Fig. 4, we can figure out that $|width-height|$ values of different face regions are always greater than 0, especially for eyes areas, mouth areas and eyes to nose areas. It increases the challenges of the emotion recognition process. Therefore, data preprocessing approaches should to be chosen carefully, which is discussed in Sec 4.1 in detail.

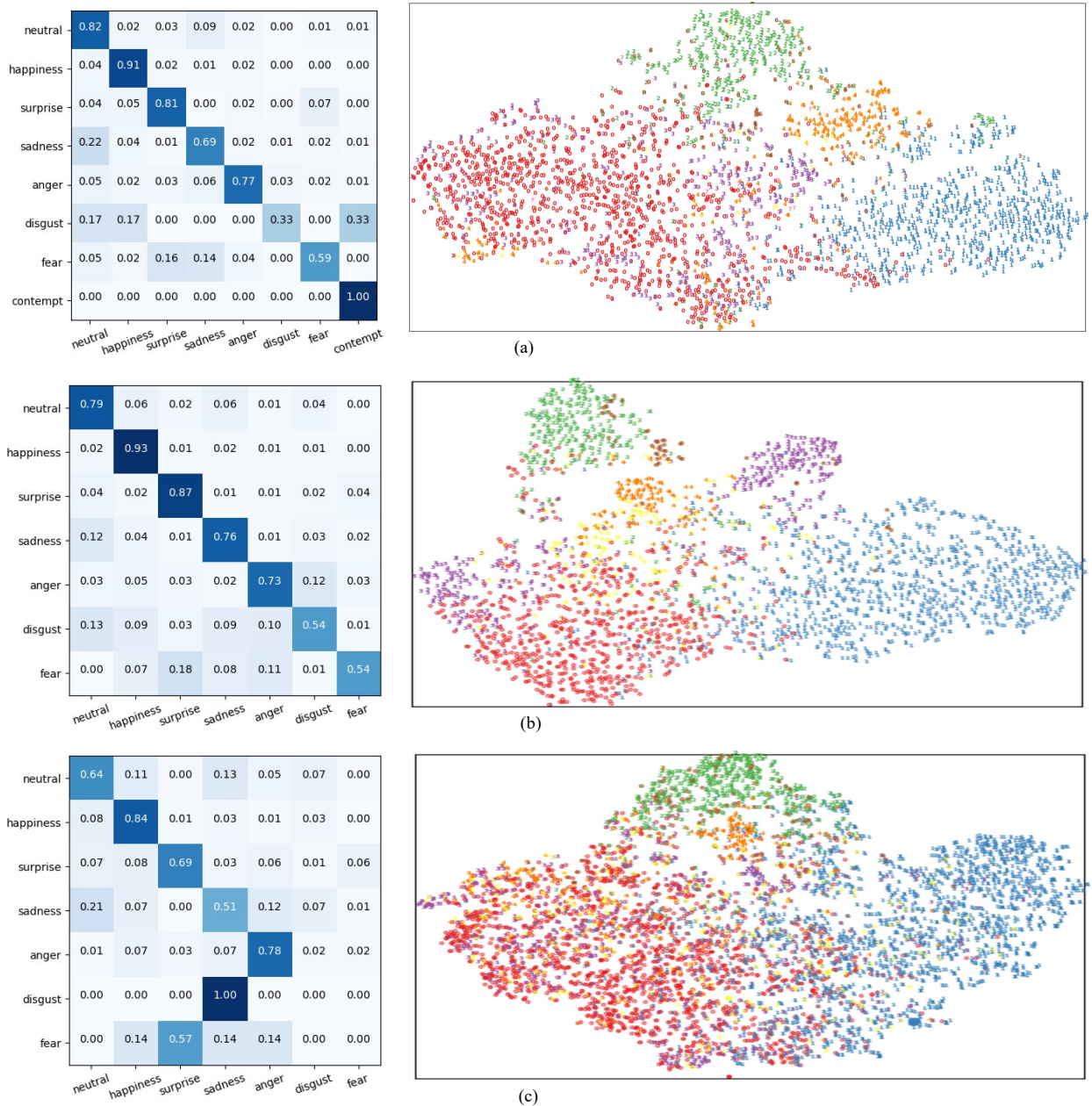

Fig. 5. (a) The performance of the FER+ training dataset based on the whole faces. Left: Visualization the confusion matrix; Right: Visualization bottleneck features through t-SNE. (b) The performance of the FER+ testing dataset based on the whole faces. Left: Visualization the confusion matrix; Right: Visualization bottleneck features through t-SNE. [0(red): neutral, 1(blue): happiness, 2(green): surprise, 3(purple): sadness, 4(orange): anger, 5(yellow): disgust, 6(brown): fear, 7(pink): contempt]

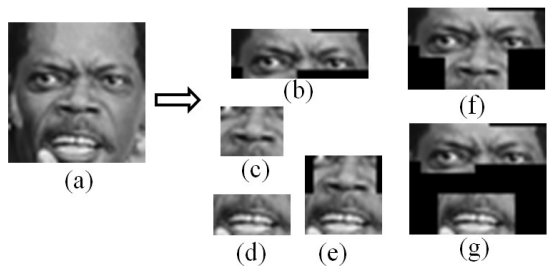

Fig.4. The whole face (a) is divided into six face regions according to landmarks, including (b) eyes areas, (c) nose areas, (d) mouth areas, (e) nose to mouth areas, (f) nose to eyes areas and (g) mouth to eyes areas.

## 4 Evaluation results

In this section, we analyze the contribution of different face areas on three datasets: FER+, RAF-DB and ExpW.

Firstly, we compare two data pre-processing methods, and choose the training approach and the testing approach for both classification model in Sec 2.1 and visualization model in Sec 2.2. Secondly, we analyze classification performance based on the whole face, which is calculated by the classification model in Sec 2.1. We treat it as a comparison experiment. Thirdly, we visualize activation parts of the inputs through CAM technique, which is based on the visualization model in Sec 2.2. Finally, we compare the generation performance of models trained by seven face areas through the classification accuracy and the confusion matrix, and we also analyze the contribution of different face areas to different emotions.

### 4.1 Training and testing approach

To recognize emotions and visualize activation parts, we train the system for both the visualization model and the classification model.

In the training process, Adam [42] optimizer is utilized to minimize the cross entropy loss. Learning rate is set to be 0.05 at first. If the classification accuracy of the testing dataset is decreased, a smaller learning rate will be utilized. As for the data augmentation methods, random crop original inputs into squares and random horizontal flip are chosen to obtain more robust emotion classifiers. Then, each image is normalized to have the same mean and variance in each channel. The maximum training epoch is set to be 100 and early stopping is applied to alleviate overfitting problem. To alleviate the randomness in the training process, we train the system five times and choose the best model according to the performance on the testing set.

In the testing process, data augmentation methods in the training process are replaced. Each face is cropped around the center.

As different images have different width and height, their aspect ratio is distinct, especially for cropped face regions such as mouth areas, whose $|width-height|$ is big. If those images are cropped into square in the data argumentation process, they will lose much information. Therefore, we utilize a padding approach to convert original inputs into standard images whose $|width-height|$ are 0.

To verify the effectiveness of the padding process, we compare classification accuracy on the RAF-DB dataset on two conditions: with padding and without padding. Classification accuracy can be found in Table 5.

Table 5   Classification accuracy (in %) of the RAF-DB testing dataset at two conditions: with padding and without padding.

| Face areas | Non-padding | Padding |
|---|---|---|
| Mouth | 55.12 | **60.07** |
| Nose | 44.43 | **49.02** |
| Eyes | 40.16 | **50.20** |
| Nose and mouth | 56.68 | **63.14** |
| Nose and eyes | 58.41 | **58.87** |
| Mouth and eyes | 67.63 | **67.83** |
| The whole faces | 77.31 | **82.69** |

Through analysis on Table 5, we can figure out that the padding approach is useful to train a better system. Therefore, we will utilize the padding approach in the following experiments.

### 4.2 Performance of the whole faces

To analyze the generation ability of trained models, we visualize the classification probabilities of the testing dataset through the confusion matrix, which is generated by the classification model in Sec 2.1. Furthermore, we treat outputs of GAP in the visualization model in Sec 2.2 as bottleneck features. And, we visualize these features through t-SNE [43], which is realized under scikit [44]. The confusion matrixes and t-SNE results are shown in Fig. 5. The same phenomenon can be found through two analysis methods.

We can figure out that the recognition performance of disgust and fear are bad through the confusion matrix in Fig. 5 (c). Besides, we can figure out that different emotion categories have a large overlap with others in t-SNE results. However, the confusion matrix and t-SNE results in Fig. 5 (a) and (b) are better than (c). Therefore, we are convinced that models trained by the FER+ dataset and the RAF-DB dataset have better generation performance than the ExpW dataset, which is related to the quality of labeling approach of each dataset.

Through further analysis on three confusion matrixes in Fig. 5, we can find that happiness always has the highest classification accuracy. Besides happiness, anger, sadness, surprise and neutral also have good performance. However, fear and disgust always have worse performance than other emotions. Such phenomenon is related with unbalanced label distribution. In the meantime, it is also related with the definition of each emotion. The definition of happiness is clear and definite for most people. However, the definition of fear and disgust are blurring. Different people can mistake fear and disgust for different emotions. In the FER+ database, fear is easily confused with surprised and sadness, and disgust is easily confused with neutral and happiness. As for the RAF-DB dataset, the definition of fear and surprise are blurring. And, the disgust is easily mistaken for neutral and anger. In the ExpW dataset, disgust is easily confused with sadness and fear is easily confused with surprise.

Through analysis on three different datasets, we can find that fear and surprise are always easily confused with each other, which is related with the human perceptions of fear and surprise. The borderline of fear and surprise are quite blurring. Some emotions contain both fear and surprise, such as frighten. Through Plutchik's three-dimensional emotion model in [45], we can also find the same phenomenon that fear and surprise are close to each other in the psychological aspect.

As contempt only appears in the FER+ dataset and only few samples are labeled into contempt, results of contempt is lack of confidence. Therefore, contempt is ignored in the following experiments.

### 4.3 CAM visualization

We visualize facial activation areas through CAM for CNNs with GAP. Heatmap is visualized through COLORMAP_JET color mapping realized under opencv, which varies from blue (low range) to green (mid range) to red (upper range). To show heatmaps on original images, we combine them together through weighted coefficient in [26]:

$$result = heatmap \times 0.4 + image \times 0.5 \quad (3)$$

Heatmaps of different emotions in three datasets are shown in Fig. 6.

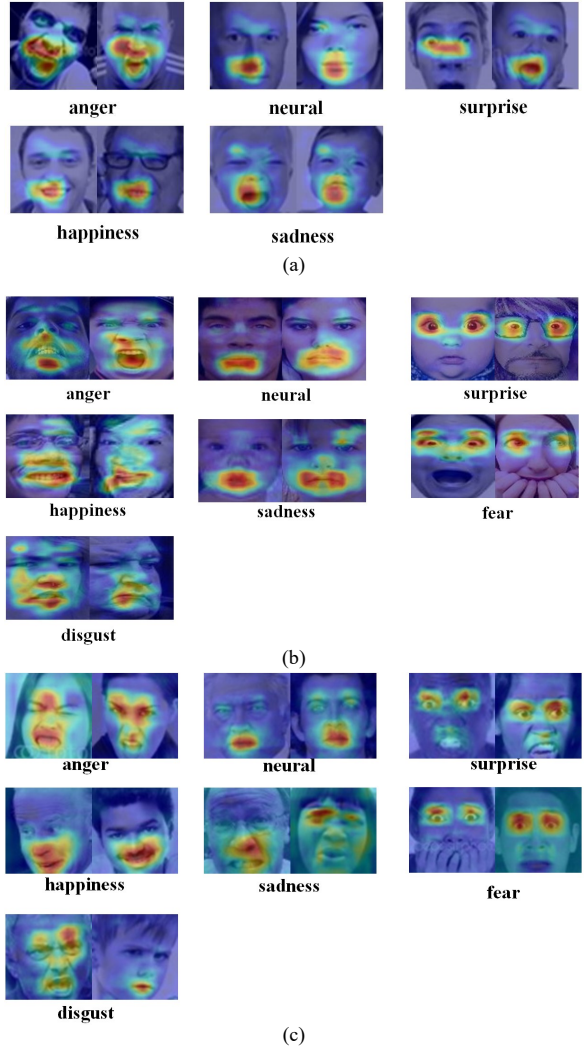

Fig.6. Heatmaps of different emotions. (a) Results on the FER+ Database. (b) Results on the RAF-DB Database. (c) Results on the ExpW Database.

As mouth areas, nose areas and eye areas are colored in most cases in Fig. 6 (a), (b) and (c), we can infer that those areas are related with emotion expression. In the meantime, there are few colors on the forehead and cheek, which shows that the forehead and cheek have less contribution to the emotion expression.

As the red refers higher activation values than the blue, we can further figure out that mouth areas convey more emotional information than nose areas and eye areas, as heatmaps for mouth regions are close to red. Besides, areas around eyes and noses are colored blue. Therefore, those areas also count.

As for happiness, mouth areas are always colored in red, and eyes areas and nose areas are always colored in blue on three datasets. It shows that happiness is related with mouth areas in most cases. And eyes areas and nose areas contribute less than mouth areas for the happiness.

As eyes areas and mouth areas are always colored in red for fear (and surprise), which shows that eyes areas and mouth areas convey more information than nose areas in those emotion states.

As for disgust and angry, those emotions are relate with the whole faces. Multiple face regions contribute to generate disgust and angry.

### 4.4 Contribution of different face areas

To compare the impact of different face areas, we train seven classification models based on seven face areas, including nose areas, mouth areas, eyes areas, nose to mouth areas, nose to eyes areas, mouth to eyes areas and the whole face areas. The training process and the testing process followed with Sec. 4.1 except inputs are replaced by corresponding face regions. For example, as for mouth areas, we only utilize mouth for both training and testing, and corresponding emotion labels are treated as outputs.

Our experiments are conducted on three datasets. Classification performance for seven face areas in the testing set is shown in Table 6. We also visualize the confusion matrix of the testing set for mouth, nose and eyes areas in Fig. 7.

Table 6 Classification accuracy (%) of different face areas in the testing set.

| Face areas | FER+ | RAD-DB | ExpW |
|---|---|---|---|
| Mouth | 73.17 | 60.07 | 64.53 |
| Nose | 68.81 | 49.02 | 53.51 |
| Eyes | 55.21 | 50.20 | 56.48 |
| Nose and mouth | 77.01 | 63.14 | 66.96 |
| Nose and eyes | 77.29 | 58.87 | 61.46 |
| Mouth and eyes | 81.57 | 67.83 | 67.27 |
| The whole faces | **81.93** | **82.69** | **71.90** |

Through analysis on Table 6 and Fig. 7, we can figure out that compound regions (such as nose and mouth areas) always have better classification performance than corresponding basic areas (such as nose areas or mouth areas). In the meantime, the whole face areas always achieve the highest classification accuracy among seven face regions. Therefore, it is reasonable to conclude that the expression approach for each emotion is related to the whole faces. And taking into account larger face areas is helpful to judge expressions more precisely.

As mouth areas have the highest classification accuracy among basic areas through analysis on Table 6, we can figure out that the mouth areas convey a lot of information about facial emotions.

As for the neutral, happiness and anger, mouth areas always gain the highest classification accuracy among basic areas through analysis on Fig. 7. It is reasonable to conclude that expression approaches for those emotions are related with mouth areas.

As for surprise, eyes areas lead to the least confusion among basic areas in most cases. We can figure out that eyes areas contain much information for the surprise.

## 5 Conclusions

Facial expression recognition is an essential aspect in the human-machine interaction. However, it faces many challenges in real-world conditions, such as illumination changes, large pose variations and partial or full occlusions, which cause different face areas with different sharpness and completeness. Therefore, we focus on answering the emotion recognition confidence based on partial faces through analysis on the contribution of different face areas to different emotions, including nose areas, mouth areas, eyes areas, nose to mouth areas, nose to eyes areas, mouth to eyes areas and the whole faces.

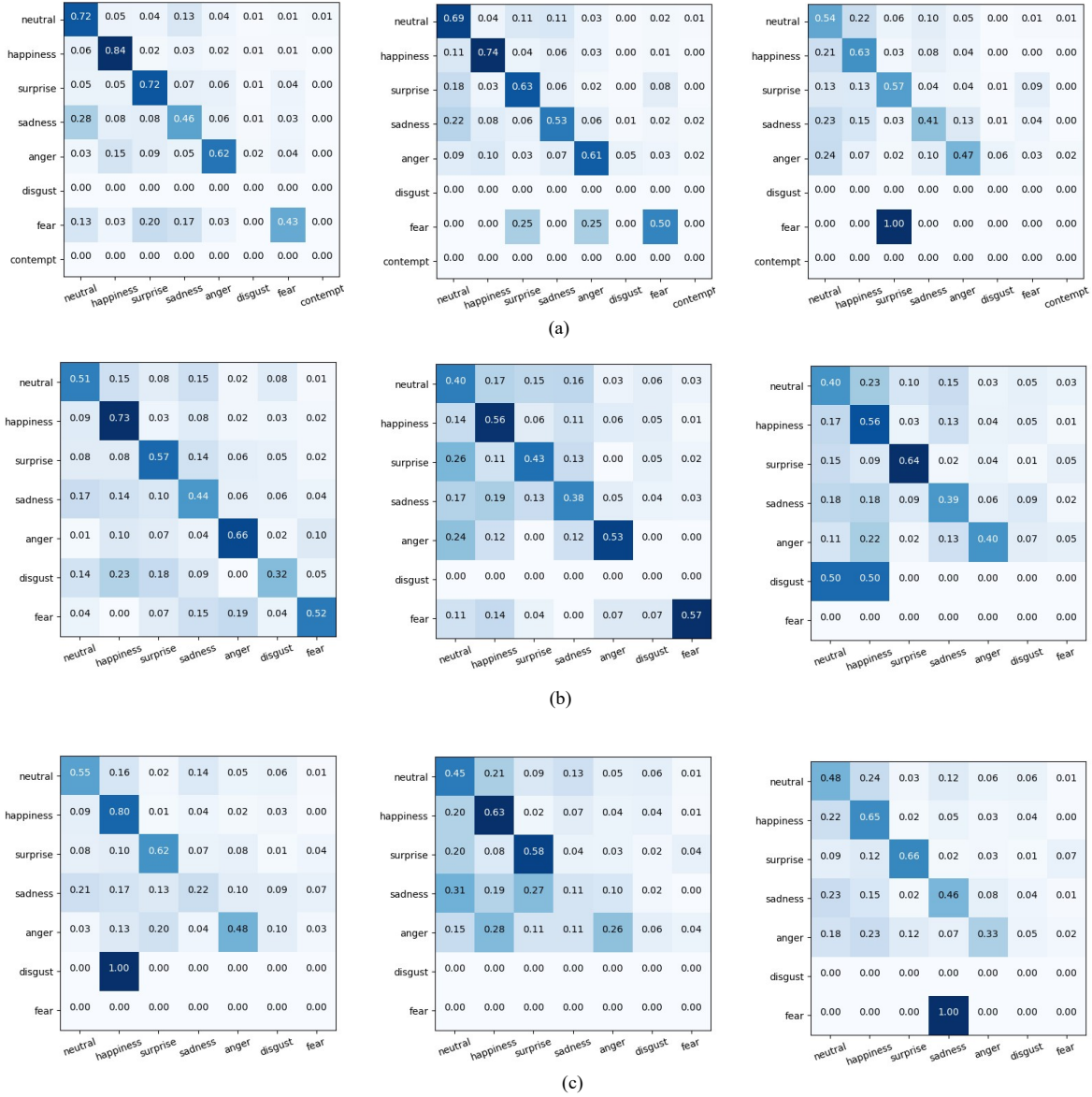

Fig. 7. (a) The performance of the FER+ testing dataset: The confusion matrix based on mouth areas (the left column), nose areas (the middle column) and eyes areas (the right column). (b) The performance of the RAF-DB testing dataset: The confusion matrix based on mouth areas (the left column), nose areas (the middle column) and eyes areas (the right column). (c) The performance of the ExpW testing dataset: The confusion matrix based on mouth areas (the left column), nose areas (the middle column) and eyes areas (the right column).

Through analysis the confusion matrix, CAM results and the classification accuracy on three different datasets (including the FER+ dataset, the RAF-DB dataset and the ExpW dataset), we can figure out universal approaches for different emotion expression. We figure out that mouth areas convey a lot of information about facial emotions, especially for the neutral, happiness and anger. And eyes areas contain much information for the surprise. Furthermore, we can judge expressions more precisely through considering larger face areas.

Our work can be combined with findings in the psychological aspect. The contribution of this paper is critical to the study of human behaviors, and it can also promote the understanding of emotion expression.

# Acknowledgments

This work is supported by the National Key Research & Development Plan of China (No. 2017YFB1002804), and the National Natural Science Foundation of China (NSFC) (NO.61425017, No.61773379, No.61332017, No.61603390, No.61771472) and the Major Program for the 325 National Social Science Fund of China (13&ZD189).

# References

[1] H. Prendinger, J. Mori, and M. Ishizuka, "Using human physiology to evaluate subtle expressivity of a virtual quizmaster in a mathematical game," in International journal of human-computer studies, pp. 231-245, 2005.

[2] B. Martinovski and D. Traum, "Breakdown in human-machine interaction: the error is the clue," in Proceedings of the ISCA tutorial and research workshop on Error handling in dialogue systems, pp. 11-16, 2003.

[3] N. Asghar, P. Poupart, J. Hoey, X. Jiang, and L. Mou, "Affective Neural Response Generation," in European Conference on Information Retrieval, pp. 154-166, 2017.

[4] H. Zhou, M. Huang, T. Zhang, X. Zhu, and B. Liu, "Emotional Chatting Machine: Emotional Conversation Generation with Internal and External Memory," in Thirty-Second AAAI Conference on Artificial Intelligence, 2018.

[5] S. Ghosh, M. Chollet, E. Laksana, L.-P. Morency, and S. Scherer, "Affect-LM: A Neural Language Model for Customizable Affective Text Generation," in ACL, 2017.

[6] N. Dalal and B. Triggs, "Histograms of oriented gradients for human detection," in IEEE Computer Society Conference on Computer Vision and Pattern Recognition, vol. 1, pp. 886-893: IEEE, 2005.

[7] T. Ojala, M. Pietikainen, and T. Maenpaa, "Multiresolution gray-scale and rotation invariant texture classification with local binary patterns," IEEE Transactions on pattern analysis and machine intelligence, vol. 24, no. 7, pp. 971-987, 2002.

[8] V. Ojansivu and J. Heikkilä, "Blur insensitive texture classification using local phase quantization," in International conference on image and signal processing, pp. 236-243: Springer, 2008.

[9] D. G. Lowe, "Distinctive image features from scale-invariant keypoints," in International journal of computer vision, vol. 60, no. 2, pp. 91-110, 2004.

[10] Y. Chen, J. Li, H. Xiao, X. Jin, S. Yan, and J. Feng, "Dual Path Networks," in Advances in Neural Information Processing Systems, pp. 4467-4475, 2017.

[11] A. Vaswani et al., "Attention is all you need," in Advances in Neural Information Processing Systems, pp. 6000-6010, 2017.

[12] L. Shen, Z. Lin, and Q. Huang, "Relay Backpropagation for Effective Learning of Deep Convolutional Neural Networks," in European conference on computer vision, pp. 467-482, 2016.

[13] L. Chen et al., "SCA-CNN: Spatial and Channel-wise Attention in Convolutional Networks for Image Captioning," in Proceedings of the IEEE Conference on Computer Vision and Pattern Recognition, pp. 5659-5667, 2017.

[14] A. Van Den Oord et al., "Wavenet: A generative model for raw audio," arXiv preprint arXiv:1609.03499, 2016.

[15] A. Krizhevsky, I. Sutskever, and G. E. Hinton, "ImageNet classification with deep convolutional neural networks," in International Conference on Neural Information Processing Systems, pp. 1097-1105, 2012.

[16] K. Simonyan and A. Zisserman, "Very Deep Convolutional Networks for Large-Scale Image Recognition," in Computer Science, 2014.

[17] C. Szegedy et al., "Going deeper with convolutions," in Proceedings of the IEEE Conference on Computer Vision and Pattern Recognition, pp. 1-9, 2015.

[18] K. He, X. Zhang, S. Ren, and J. Sun, "Deep Residual Learning for Image Recognition," in IEEE Conference on Computer Vision and Pattern Recognition, pp. 770-778, 2016.

[19] G. Huang, Z. Liu, L. V. D. Maaten, and K. Q. Weinberger, "Densely Connected Convolutional Networks," in IEEE Conference on Computer Vision and Pattern Recognition, pp. 2261-2269, 2017.

[20] P. Ekman and W. Friesen, "The facial action coding system (FACS): a technique for the measurement of facial action Vol," in Consulting Psychologists. Palo Alto, CA, 1978.

[21] Y. L. Tian, T. Kanade, and J. F. Cohn, "Recognizing Action Units for Facial Expression Analysis," in IEEE Trans Pattern Anal Mach Intell, vol. 23, no. 2, pp. 97-115, 2001.

[22] L. Wang, R.-F. Li, K. Wang, and J. Chen, "Feature representation for facial expression recognition based on FACS and LBP," in International Journal of Automation and Computing, vol. 11, no. 5, pp. 459-468, 2014.

[23] X. Sun, M. Lv, C. Quan, and F. Ren, "Improved Facial Expression Recognition Method Based on ROI Deep Convolutional Neural Network," in International Conference on Affewctive Computing and Intelligent Interaction (ACII), 2017, pp. 256-261.

[24] Z. Wei, Y. Zhang, M. Lin, J. Guan, and S. Gong, "Multimodal learning for facial expression recognition," in Pattern Recognition, vol. 48, no. 10, pp. 3191-3202, 2015.

[25] F. K. Zaman, A. A. Shafie, and Y. M. Mustafah, "Robust face recognition against expressions and partial occlusions," in International Journal of Automation & Computing, vol. 13, no. 4, pp. 319-337, 2016.

[26] B. Zhou, A. Khosla, A. Lapedriza, A. Oliva, and A. Torralba, "Learning deep features for discriminative localization," in Proceedings of the IEEE Conference on Computer Vision and Pattern Recognition, 2016, pp. 2921-2929.

[27] C. Busso et al., "Analysis of emotion recognition using facial expressions, speech and multimodal information," in Proceedings of the 6th international conference on Multimodal interfaces, vol. 38, no. 4, pp. 205-211, 2004.

[28] S. Ioffe and C. Szegedy, "Batch normalization: Accelerating deep network training by reducing internal covariate shift," in International conference on machine learning, pp. 448-456, 2015.

[29] K. Jarrett, K. Kavukcuoglu, M. A. Ranzato, and Y. Lecun, "What is the best multi-stage architecture for object recognition?," in IEEE International Conference on Computer Vision, pp. 2146-2153, 2010.

[30] E. Barsoum, C. Zhang, C. C. Ferrer, and Z. Zhang, "Training deep networks for facial expression recognition with crowd-sourced label distribution," in ACM International Conference on Multimodal Interaction, pp. 279-283, 2016.

[31] S. Li, W. Deng, and J. P. Du, "Reliable Crowdsourcing and Deep Locality-Preserving Learning for Expression Recognition in the Wild," in IEEE Conference on Computer Vision and Pattern Recognition, pp. 2584-2593, 2017.

[32] Z. Zhang, P. Luo, C. L. Chen, and X. Tang, "From Facial Expression Recognition to Interpersonal Relation Prediction," in International Journal of Computer Vision, vol. 126, no. 5, pp. 550-569, 2016.

[33] M. J. Lyons, J. Budynek, and S. Akamatsu, "Automatic classification of single facial images," in Pattern Analysis & Machine Intelligence IEEE Transactions on, vol. 21, no. 12, pp. 1357-1362, 1999.

[34] M. Pantic, M. Valstar, R. Rademaker, and L. Maat, "Web-based database for facial expression analysis," in Proceedings of the IEEE International Conference on Multimedia and Expo, p. 5-pp, 2005.

[35] G. Zhao, X. Huang, M. Taini, and S. Z. Li, "Facial expression recognition from near-infrared videos," in Image & Vision Computing, vol. 29, no. 9, pp. 607-619, 2011.

[36] P. Lucey, J. F. Cohn, T. Kanade, and J. Saragih, "The Extended Cohn-Kanade Dataset (CK+): A complete dataset for action unit and emotion-specified expression," in Computer Vision and Pattern Recognition Workshops, pp. 94-101, 2010.

[37] A. Dhall, O. Ramana Murthy, R. Goecke, J. Joshi, and T. Gedeon, "Video and image based emotion recognition challenges in the wild: Emotiw 2015," in Proceedings of the 2015 ACM on International Conference on Multimodal Interaction, pp. 423-426: ACM, 2015.

[38] D. E. King, "Dlib-ml: A Machine Learning Toolkit," in Journal of Machine Learning Research, vol. 10, no. 3, pp. 1755-1758, 2009.

[39] I. J. Goodfellow et al., "Challenges in representation learning: A report on three machine learning contests," in Neural Networks, vol. 64, pp. 59-63, 2015.

[40] B. Yang, J. Yan, Z. Lei, and S. Z. Li, "Aggregate channel features for multi-view face detection," in IEEE International Joint Conference on Biometrics, pp. 1-8, 2014.

[41] V. Kazemi and J. Sullivan, "One millisecond face alignment with an ensemble of regression trees," in IEEE Conference on Computer Vision and Pattern Recognition, pp. 1867-1874, 2014.

[42] D. P. Kingma and J. Ba, "Adam: A method for stochastic optimization," arXiv preprint arXiv:1412.6980, 2014.

[43] L. V. D. Maaten and G. Hinton, "Visualizing Data using t-SNE," in Journal of Machine Learning Research, vol. 9, no. 2605, pp. 2579-2605, 2008.

[44] F. Pedregosa et al., "Scikit-learn: Machine Learning in Python," in Journal of Machine Learning Research, vol. 12, no. 10, pp. 2825-2830, 2013.

[45] R. Plutchik, "The multifactor-analytic theory of emotion," in Journal of Psychology, vol. 50, no. 1, pp. 153-171, 1960.

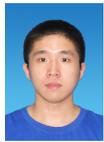
**Zheng Lian** Ph. D candidate at the National Laboratory of Pattern Recognition (NLPR), Institute of Automation, Chinese Academy of Sciences, received the B.E. degree from Beijing University of Posts and Telecommunications. His research interests cover affective computing, deep learning, multimodal emotion recognition.

E-mail: lianzheng2016@ia.ac.cn

ORCID iD: 0000-0001-9477-0599

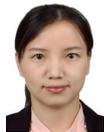
**Ya Li** received the B.E. degree from University of Science and Technology of China (USTC) in 2007, and Ph.D. degree from NLPR, CASIA in 2012. She is currently an Associate Professor in Institute of Automation, Chinese Academy of Sciences (CASIA), Beijing. Her general interests include Affective Computing and Human-Computer Interaction. She has published more than 50 papers in the related journals and conferences, such as Speech Communication, ICASSP, INTERSPEECH, and ACII. She has won the second prize of Beijing Science and Technology Award in 2014. She has also won the Best Student Paper in INTERSPEECH 2016.

E-mail: yli@nlpr.ia.ac.cn (Corresponding author)

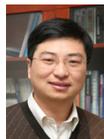
**Jianhua Tao** received PhD from Tsinghua University in 2001. He is Winner of the National Science Fund for Distinguished Young Scholars and the deputy director in NLPR, CASIA. He has directed many national projects, including "863", National Natural Science Foundation of China. His interests include speech synthesis, affective computing and pattern recognition. He has published more than eighty papers on journals and proceedings including IEEE Trans. on ASLP, and ICASSP, INTERSPEECH. He also serves as the steering committee member for IEEE Transactions on Affective Computing and the chair or program committee member for major conferences, including ICPR, Interspeech, etc.

E-mail: jhtao@nlpr.ia.ac.cn

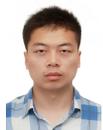
**Jian Huang** Ph. D candidate at the National Laboratory of Pattern Recognition (NLPR), Institute of Automation, Chinese Academy of Sciences, received the B.E. degree from Wuhan University. He had published the papers in INTERSPEECH and ICASSP. His research interests cover affective computing, deep learning, multimodal emotion recognition.

E-mail: jian.huang@nlpr.ia.ac.cn

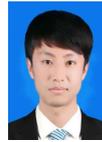
**Mingyue Niu** received his Master degree from the Department of Applied Mathematics, Northwestern Polytechnical University (NWPU), China, in 2014. Currently, he is pursuing a Ph.D. in the National Laboratory of Pattern Recognition (NLPR), Institute of Automation, Chinese Academy of Sciences (CASIA). His research interests include Affective Computing and Human-Computer Interaction.

E-mail: niumigyue2017@ia.ac.cn